\pdfoutput=1

\documentclass[11pt]{article}

\usepackage{latex/acl}

\usepackage{times}
\usepackage{latexsym}

\usepackage[T1]{fontenc}

\usepackage[utf8]{inputenc}

\usepackage{microtype}

\usepackage{inconsolata}

\usepackage{soul}
\usepackage{url}
\usepackage[utf8]{inputenc}
\usepackage{caption}
\usepackage{graphicx}
\usepackage{amsmath,amsfonts,amssymb}
\usepackage{booktabs}
\usepackage{subcaption}
\usepackage{mathtools}
\usepackage{makecell}
\usepackage{placeins} 

\usepackage{tabu}

\usepackage{enumitem}
\usepackage{siunitx}
\usepackage{multirow}
\usepackage{setspace}
\usepackage{algcompatible}
\usepackage[noend]{algpseudocode}
\makeatletter
\def\BState{\State\hskip-\ALG@thistlm}
\makeatother

\algdef{SE}[SUBALG]{Indent}{EndIndent}{}{\algorithmicend\ } 
\algtext*{Indent}
\algtext*{EndIndent}

\usepackage{xcolor} 
\newcommand{\parm}{\mathord{\color{black!33}\bullet}}

\usepackage{wrapfig} 
\usepackage{colortbl} 

\newcommand\MULTIFCDS{MultiFC~}
\newcommand\SNOPESDS{Snopes~}

\title{Fact Checking Beyond Training Set}

\author{Payam Karisani \\
  UIUC \\
  \texttt{karisani@illinois.edu} \\\And
  Heng Ji \\
  UIUC \\
  \texttt{hengji@illinois.edu} \\}
  
\begin{document}
\maketitle
\begin{abstract}

Evaluating the veracity of everyday claims is time consuming and in some cases requires domain expertise. We empirically demonstrate that the commonly used fact checking pipeline, known as the retriever-reader, suffers from performance deterioration when it is trained on the labeled data from one domain and used in another domain. Afterwards, we delve into each component of the pipeline and propose novel algorithms to address this problem. We propose an adversarial algorithm to make the retriever component robust against distribution shift. Our core idea is to initially train a bi-encoder on the labeled source data, and then, to adversarially train two separate document and claim encoders using unlabeled target data. We then focus on the reader component and propose to train it such that it is insensitive towards the order of claims and evidence documents. Our empirical evaluations support the hypothesis that such a reader shows a higher robustness against distribution shift. To our knowledge, there is no publicly available multi-topic fact checking dataset. Thus, we propose a simple automatic method to re-purpose two well-known fact checking datasets. We then construct eight fact checking scenarios from these datasets, and compare our model to a set of strong baseline models, including recent domain adaptation models that use GPT4 for generating synthetic data.\footnote{\raggedright Code and data are available at https://github.com/p-karisani/OODFC}

\end{abstract}

\section{Introduction} \label{sec:intro}

\begin{figure}
    \centering
    \begin{subfigure}[t]{1.0\linewidth}
        \centering
        \includegraphics[width=\linewidth]{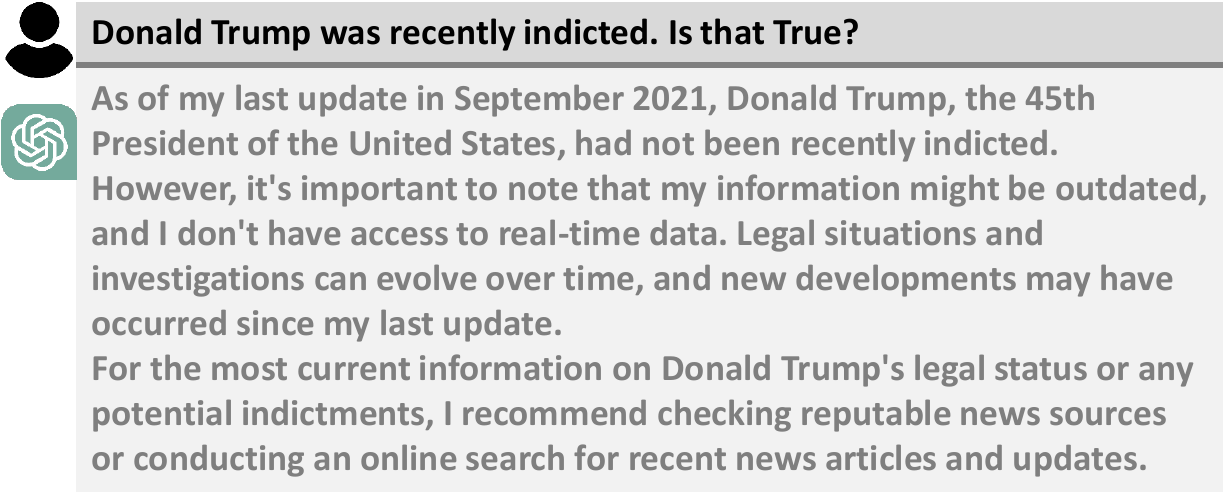}
        \caption{ChatGPT Response}
        \label{fig:chatgpt-response}
    \end{subfigure} \\
    \begin{subfigure}[t]{1.0\linewidth}
        \centering
        \includegraphics[width=\linewidth]{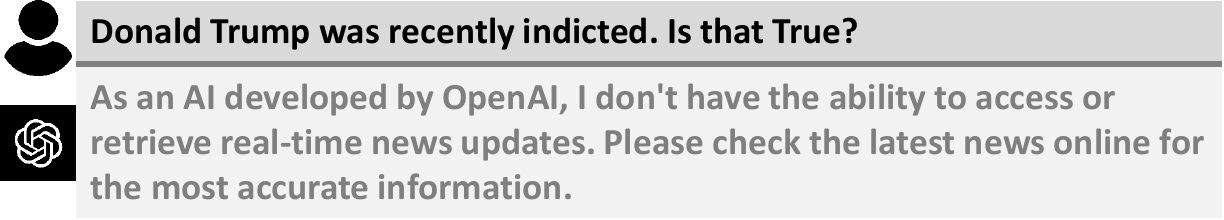}
        \caption{GPT 4 Response}
        \label{fig:gpt4-response}
    \end{subfigure}
    \caption{On August 2, 2023, the Associated Press (and other news outlets) reported that Donald Trump was indicted. The questions were asked from the LLMs on August 17, 2023. As of December 2023, these models are still unable to verify this claim.}
\end{figure}

Fact checking is the process of applying a veracity rating to a particular statement or claim \citep{fc-first}. Automatic fact checking is beneficial for curbing misinformation, and also for supporting professional human fact-checkers \citep{fc-survey}. Due to the growing size of the Web, this task is increasingly becoming more challenging. The performance of fact checking systems depends on the availability of evidence resources, and this makes the large commercial language models, in many cases, unsuitable for the task. We demonstrate this by reporting an experiment in Figures \ref{fig:chatgpt-response} and \ref{fig:gpt4-response}. We see that two existing large language models, i.e., ChatGPT and GPT 4, are unable to verify a simple claim regarding the former president of the US. At the time of carrying out this experiment, the indictment of Donald Trump was widely being discussed on news outlets, such as the Associate Press,\footnote{\raggedright Available~at: https://apnews.com/article/trump-indicted-jan-6-investigation-special-counsel-debb59bb7a4d9f93f7e2dace01feccdc} and the social media websites. Another factor that contributes to the difficulty of the task is the technological requirements. More specifically, existing automated fact checking systems rely on a pipeline of components to retrieve evidence documents and to infer the final verdict \citep{fc-survey}. As stated by \citet{fc-survey-new}, coordinating the components within such a pipeline presents an extra challenge.

Given these challenges, it is desirable to know how much a trained fact checking pipeline generalizes across domains. To our knowledge, little to no work has been done to investigate this area. There exist a few studies \citep{multi-fc-ds,fc-scientific,fc-multi-ling} that report experiments on the transferability of the fact checking pipeline across various platforms, e.g., from Wikipedia to scientific repositories, or from a fact checking website to another one. While these are valuable observations, these platforms, e.g., Wikipedia and the scientific repositories, have shared domains and topics.\footnote{We borrow the word ``domain'' in a broad sense, and the word ``topic'' in a narrow sense from the domain adaptation literature \citep{dom-ada-first}. In the domain adaptation literature ``domain'' (or ``topic'') refers to a particular genre (or sub-genre) of text, such as politics, business, or arts.} This can potentially distort certain reported conclusions, particularly, the conclusions on the generalizability of the pipeline to unseen scenarios. Additionally, the solution proposed by these studies to enhance generalization is very limited. They primarily rely on pretraining the pipeline in one domain, and then, using it in another domain.

In the next section, we provide a background on the fact checking pipeline. We then report a case study to show that a pipeline trained on out-of-domain data is not as competitive as the one trained on in-domain data. We continue our study by focusing on the two primary components of the pipeline, i.e., the retriever and the reader, and propose two novel algorithms to enhance their performance. Particularly, we use a bi-encoder dense retrieval model as the retriever, and propose an adversarial algorithm to enhance its robustness under distribution shift. We then exploit a previously unknown weakness of language models in detecting the reversal relationship between input statements, and propose an augmentation algorithm to provide the reader with more cues.

To evaluate our pipeline, we use a public API to re-purpose the Snopes \citep{snopes-ds} and the MultiFC \citep{multi-fc-ds} fact checking datasets. We extract eight fact checking scenarios out of these two datasets, and compare our proposed components individually to the state-of-the-art domain adaptation techniques, including the ones that exploit GPT~4. We also demonstrate that our entire fact checking pipeline outperforms the alternative pipelines that use these techniques. In summary, our contributions are threefold:
\begin{itemize}[noitemsep,topsep=0pt,leftmargin=*]
    \item We propose a method for the claim retriever under domain shift. Our method is novel and unprecedented. We empirically show that it outperforms existing domain adaptation models.
    \item We exploit the weakness of language models in detecting the reversal relationship in input data and propose to train the reader such that it is insensitive to the order of claim and evidence documents. This helps the reader to extract more cues from the data. Our finding about language models as well as our algorithm to partially resolve the issue are novel and unprecedented.
    \item We compare our pipeline to a set of pipelines that consist of strong domain adaptation methods. We demonstrate that ours is state of the art. 
\end{itemize}

\section{Preliminaries} \label{sec:background}

\begin{figure*}
    \centering
        \begin{subfigure}[t]{0.21\textwidth}
            \includegraphics[width=\textwidth]{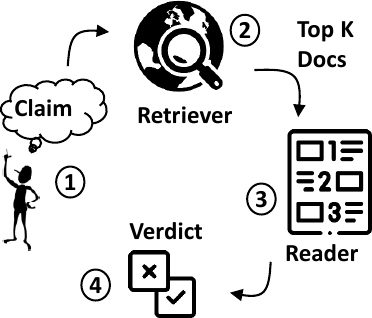}
            \caption{FC Pipeline\footnotemark{} }
            \label{fig:pipeline}
        \end{subfigure}~
        \begin{subfigure}[t]{0.21\textwidth}
            \includegraphics[width=\textwidth]{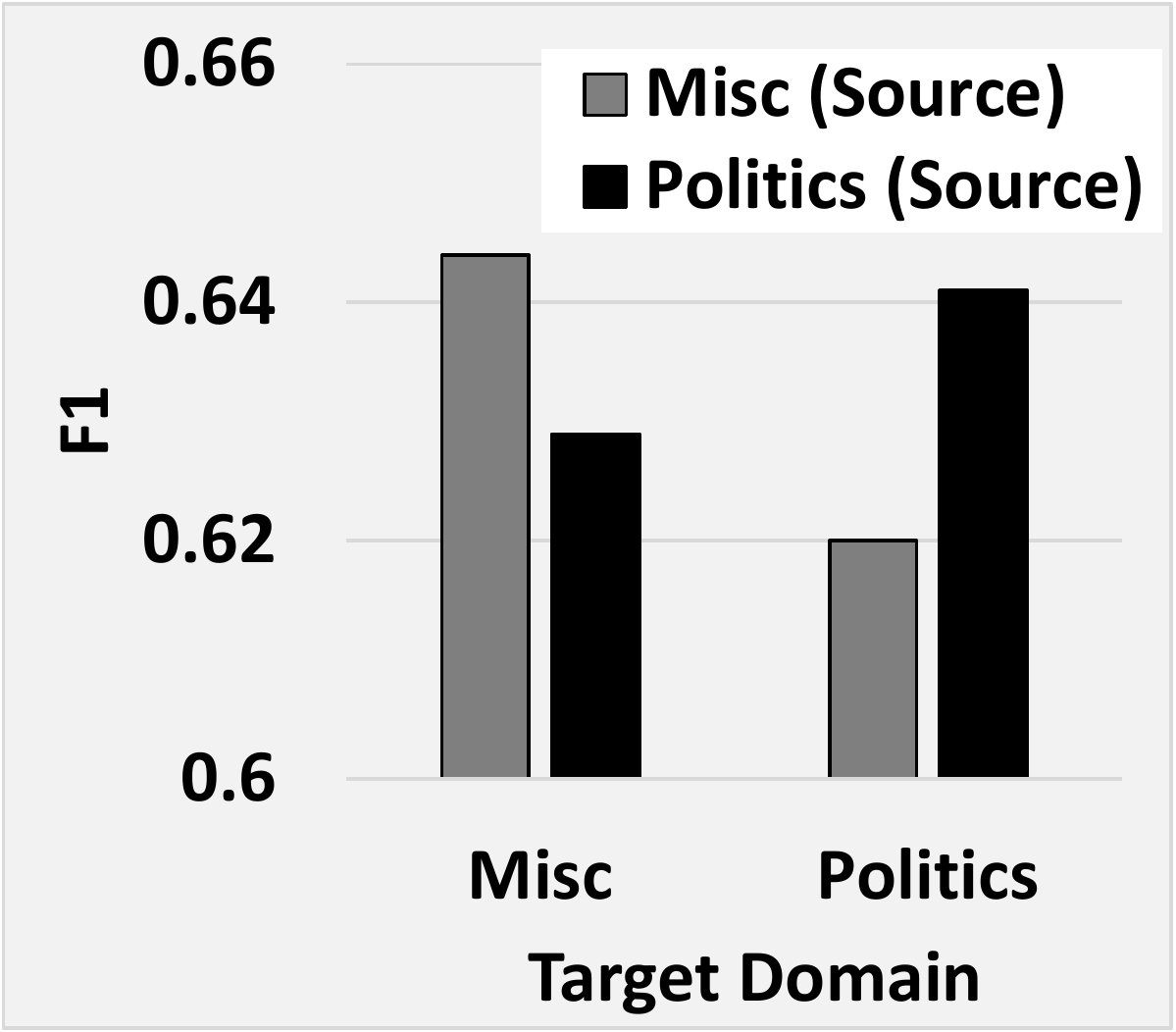}
            \caption{F1 of the Pipeline}
            \label{fig:indomain-fc-f1}
        \end{subfigure}~
        \begin{subfigure}[t]{0.21\textwidth}
            \includegraphics[width=\textwidth]{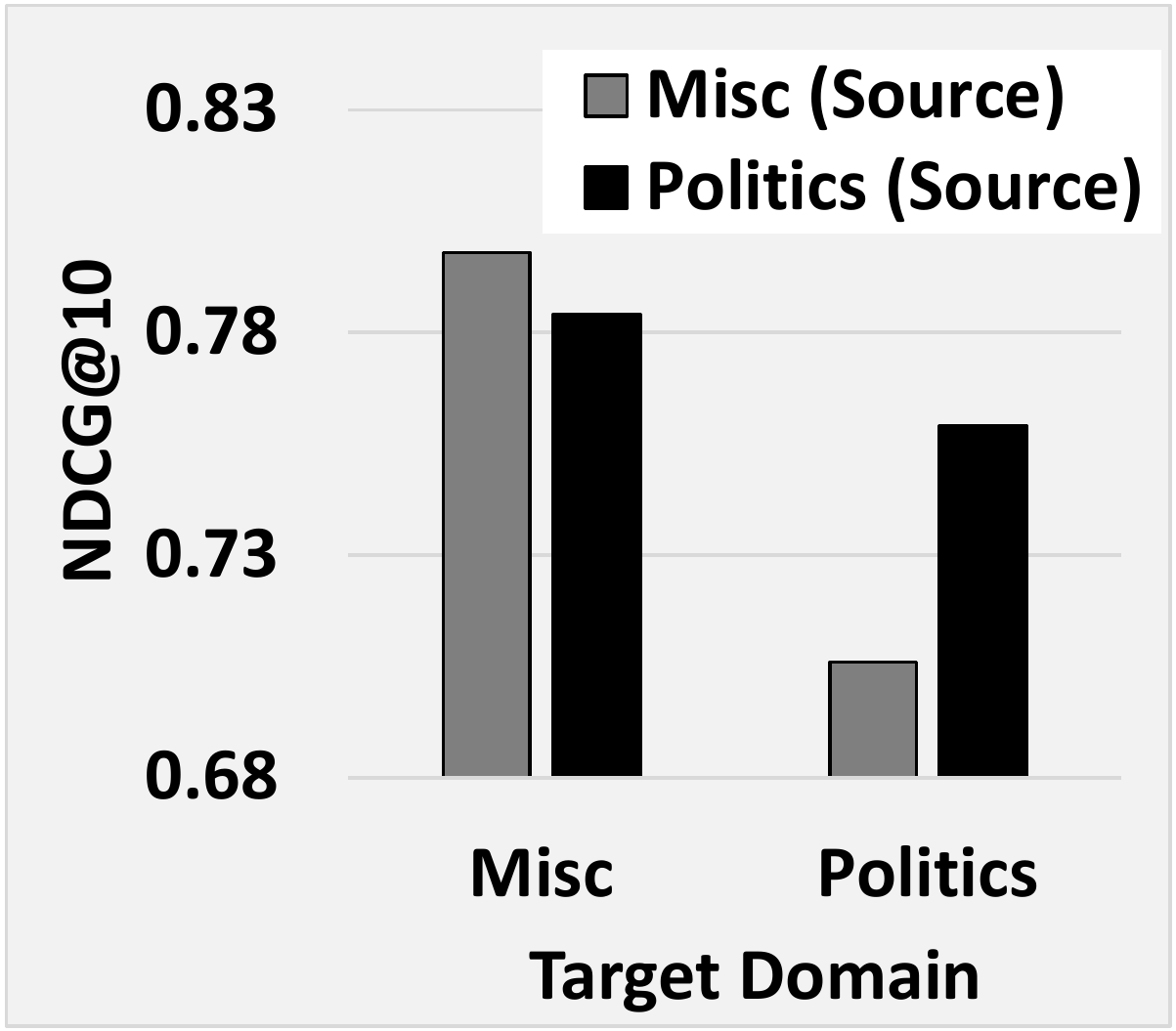}
            \caption{F1 of the Retriever}
            \label{fig:indomain-retriever-ndcg}
        \end{subfigure}~
        \begin{subfigure}[t]{0.21\textwidth}
            \includegraphics[width=\textwidth]{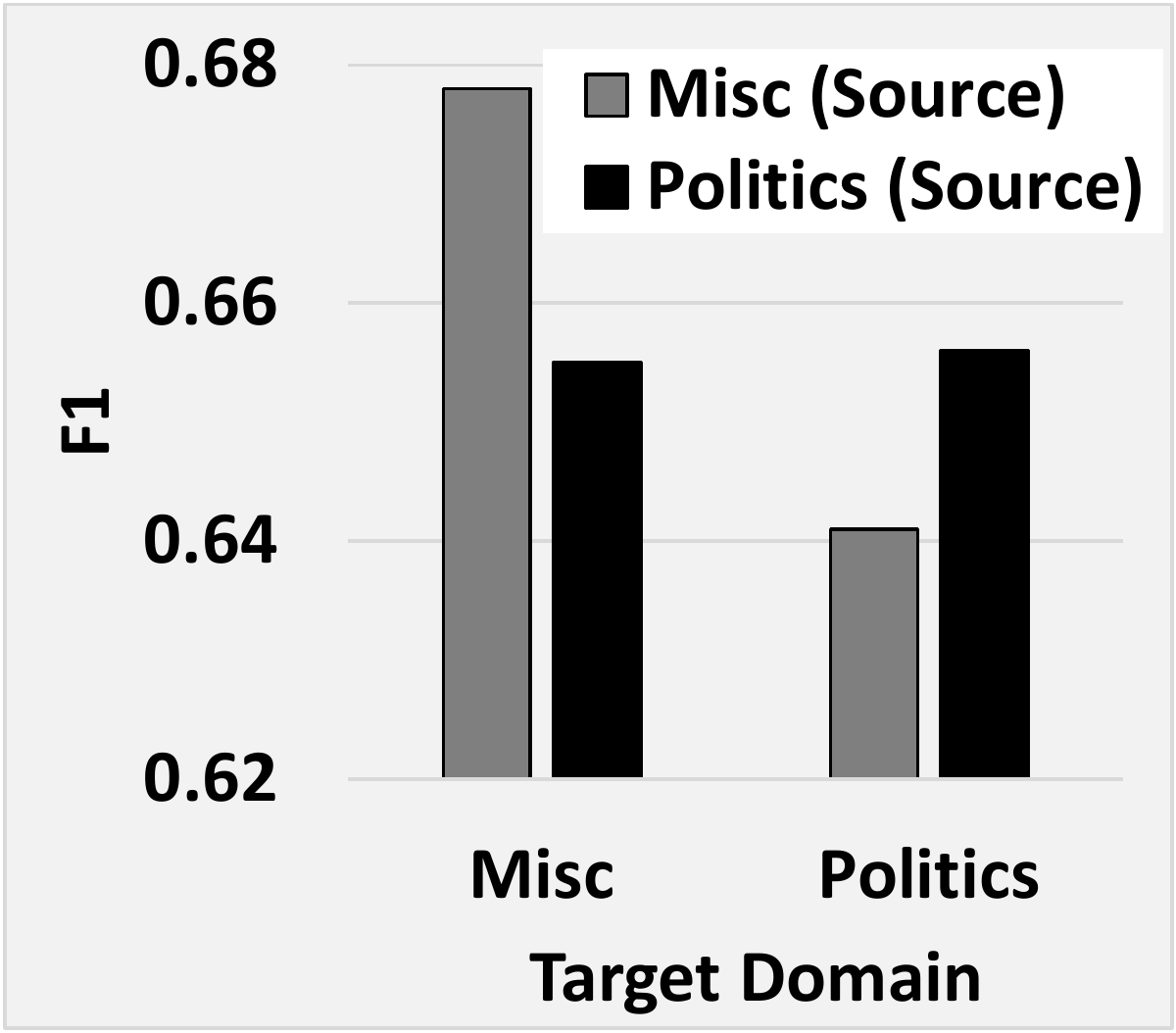}
            \caption{F1 of the Reader}
            \label{fig:indomain-reader-f1}
        \end{subfigure} 
    \caption{\textbf{\ref{fig:pipeline})} Commonly used fact checking (FC) pipeline consists of a retrieval model (called the retriever), and a veracity prediction model (called the reader). \textbf{\ref{fig:indomain-fc-f1})} The performance (Macro F1) of the pipeline across two domains (Misc vs Politics) in two scenarios (in-domain vs out-of-domain). \textbf{\ref{fig:indomain-retriever-ndcg})} The performance (NDCG@10) of the retriever across the two mentioned domains. \textbf{\ref{fig:indomain-reader-f1})} The performance (Macro F1) of the reader across the two domains.}
\end{figure*}
\footnotetext{The icons used in the figure have been downloaded from www.flaticon.com.}

\textbf{Background.} Existing fact checking systems \citep{fc-survey,fc-survey-new} primarily rely on two components: 1) a document retrieval model, called ``retriever'', and 2) a veracity prediction model, called ``reader''. See Figure \ref{fig:pipeline} for an illustration. The retriever views the input claim as a query and returns the top evidence documents that are deemed relevant to the claim---the search is usually performed over a pre-indexed corpus. 
As the reader, existing studies usually train a classifier over the concatenation of the retrieved documents and the given claim \citep{fc-survey-new}.\footnote{Depending on the architecture, practitioners may add pre-processing steps, such as rationale extraction, or post-processing steps, such as justification production. We focus on the essential components.} As stated by \citet{fc-scientific} and \citet{fc-survey}, the veracity prediction step resembles the natural language inference task (NLI). The output of the veracity prediction component can be the word ``Support'' or the word ``Refute''---depending on the system design, a third candidate output can be also added as ``Neutral''.

As it can be seen, developing, scaling up, and maintaining a fact checking system involves a lot of expertise, time, and budget. On the other hand, when such a system is deployed, even a small deterioration or improvement in performance can have profound impacts. Detecting an unsupported claim early enough, and then, taking timely actions on the media can be invaluable. Therefore, it is crucial to know if such a system is generalizable. In other words, if a model trained on the labeled data from one domain (i.e., source domain), demonstrates the same efficacy if it is used to verify the claims in another domain (i.e., target domain). In order to answer this question, below, we report a case study.

\textbf{Setup.} We compare the performance of in-domain fact checking compared to out-of-domain fact checking across two domains of ``Miscellany'' and ``Politics''.
Each domain in this experiment has 7,900 claims, and each claim has two evidence documents. In each domain, 60\% of data was used for training and 40\% for testing. The claims are labeled either as Support or Refute. 

In this experiment, the retriever is a bi-encoder \citep{dpr} pretrained using the algorithm proposed by \citet{contriever}. The reader is a RoBERTa-based
model \citep{roberta} pretrained on the SNLI and MultiNLI datasets \citep{multi-nli}. Apart from these pretraining steps, all the models are finetuned in the source domains (using the labeled data), and then, evaluated on the target domain. We assume the target domain has no labeled data during the training. The target labels are used only for evaluation. We report Macro F1 for the classification tasks and NDCG@10 for the ranking tasks.

\textbf{Observations.} Figure \ref{fig:indomain-fc-f1} reports the performance of the pipeline in the in-domain scenarios compared to the out-of-domain scenarios. We see that the performance in both of the out-of-domain scenarios (i.e., Politics$\rightarrow$Misc and Misc$\rightarrow$Politics) is worse than their in-domain counterparts. This raises the question about the root of this performance deterioration. To reveal the cause, we report the performance of each underlying component in isolation. To evaluate the performance of the reader in isolation, we assume that the retriever perfectly returns all the relevant evidence documents. Figures \ref{fig:indomain-retriever-ndcg} and \ref{fig:indomain-reader-f1} report the results. We see the same trend in both experiments. Both components suffer from distribution shift between the in-domain and out-of-domain training. 
In the next section, we formally describe the problem statement, and then, we propose solutions to enhance the performance of the pipeline.


\section{Proposed Model} \label{sec:methods}

\begin{figure*}
    \setlength\arrayrulewidth{1pt} 
    \centering
    \begingroup 
    \setlength{\tabcolsep}{1pt} 
    \begin{tabular}{c!{\color{gray}\vrule width 1pt}c!{\color{gray}\vrule width 1pt}c!{\color{gray}\vrule width 1pt}c} 
        \begin{subfigure}[t]{0.24\textwidth}
            \includegraphics[width=\textwidth]{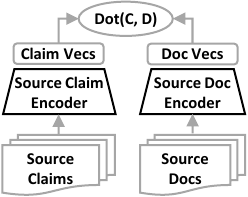}
            \caption{Training Source Model}
            \label{fig:retriever-a}
        \end{subfigure} &
        \begin{subfigure}[t]{0.24\textwidth}
            \includegraphics[width=\textwidth]{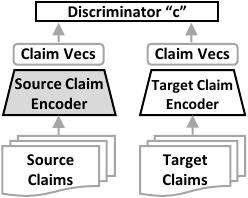}
            \caption{Target Claim Encoder}
            \label{fig:retriever-b}
        \end{subfigure} &
        \begin{subfigure}[t]{0.24\textwidth}
            \includegraphics[width=\textwidth]{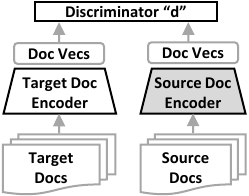}
            \caption{Target Doc Encoder}
            \label{fig:retriever-c}
        \end{subfigure} &
        \begin{subfigure}[t]{0.24\textwidth}
            \includegraphics[width=\textwidth]{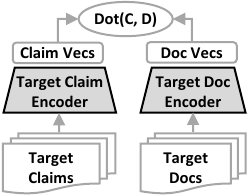}
            \caption{Testing Target Model}
            \label{fig:retriever-d}
        \end{subfigure} \\
    \end{tabular}
    \endgroup 
    \caption{\textbf{\ref{fig:retriever-a}}) The source retriever is a bi-encoder, and uses dot product as the loss function. \textbf{\ref{fig:retriever-b}}) We fix the parameters of the source claim encoder, and adversarially train the target claim encoder to mimic the source model. This step is done using unlabeled data in the two domains. \textbf{\ref{fig:retriever-c}}) Next, we fix the parameters of the source document encoder, and adversarially train the target document encoder. Similarly, this step does not need labeled data. \textbf{\ref{fig:retriever-d}}) The two trained target encoders can be used for the retrieval task in the target domain. The components that have gray outline show the inputs, outputs, and objective terms. The rest are neural networks. The parameters of the components that have gray background are fixed during training. }
\end{figure*}

\subsection{Problem Statement} 
\label{subsec:problem}

In the source domain $S$, we are given a set of labeled claims and their evidence documents denoted by ${\{(\textsc{c}_{i}^{s},y_{i}^{s},V_{i}^{s})\}}_{i=1}^{n_{s}}$, where $n_{s}$ is the number of claims in this domain, $\textsc{c}_{i}^{s}$ is the \textit{i-th} labeled claim, $y_{i}^{s}$ is the veracity of the claim---i.e., Support, Refute, or optionally Neutral---and $V_{i}^{s}$ is the set of evidence documents for supporting the assigned label. We denote the set of all the source claims by $C^s{=}{\{\textsc{c}_{i}^{s}\}}_{i=1}^{n_{s}}$, and the set of all the evidence documents by $D^s{=}{\{\textsc{d}_{j}^{s}\}}_{j=1}^{m_{s}}$, where $\textsc{d}_{j}^{s}$ is \textit{j-th} evidence document, and $m_{s}$ is the number of evidence documents in the set. Note that $V_{\parm}^{s} {\subset} D^{s}$. In the target domain $T$, we are given a set of unlabeled claims $C^t{=}{\{\textsc{c}_{i}^{t}\}}_{i=1}^{n_{t}}$, and a corpus of evidence documents~$D^t{=}{\{\textsc{d}_{j}^{t}\}}_{j=1}^{m_{t}}$. 

We opt to minimize the prediction error of the fact checking pipeline in the target domain, using the labeled data from the source domain and the unlabeled data from the target domain. Note that there is a distribution shift between the claims in the domains S and T. That is, the claims in these two domains involve distinct topics, discuss distinct entities, and refer to distinct events. 

Following existing studies \citep{fc-survey, fc-survey-new}, our model adopts the pipeline illustrated in Figure \ref{fig:pipeline}. We individually train each component using the labeled data from the source domain and the unlabeled data from the target domain. During testing, we plug the trained components into the pipeline to predict the veracity of the claims in the target domain. In the next section, we discuss our algorithm for training the retriever. We then propose our method for training the reader. We conclude the section by providing a summary of the entire training and testing procedures.

\subsection{Adversarial Training for Evidence Retrieval} \label{subsec:retriever}

We use a bi-encoder model \citep{dpr} as the retriever. This model consists of two encoders $f_{c}(\parm)$ and $f_{d}(\parm)$ to project the claims and evidence documents into low dimensional dense vectors respectively. Figure \ref{fig:retriever-a} illustrates the architecture of this model. To obtain a similarity score between a claim and evidence documents, a dot product operator is applied to the outputs of the encoders, i.e., for a given claim $\textsc{c}$ and an evidence document $\textsc{d}$ we have $sim(\textsc{c},\textsc{d}){=}{f_{c}(\textsc{c})}^{\intercal}\cdot{f_{d}(\textsc{d})}$.

To train this model in the source domain $S$, where labeled data is available, we can use the relevant evidence documents as positive examples, and the irrelevant evidence documents as negative examples. Then, we can minimize the negative log-likelihood loss term as follows:
\begin{equation}
\small
\setlength{\jot}{0pt}
\setlength{\abovedisplayskip}{0pt}
\setlength{\belowdisplayskip}{0pt}
\medmuskip=0mu
\thinmuskip=0mu
\thickmuskip=0mu
\nulldelimiterspace=0pt
\scriptspace=0pt
\begin{split}
\mathcal{L}_{f^s}= &\sum_{i=1}^{n_s} -\log \\
& \frac{\exp{(sim(\textsc{c}_{i}^{s},\textsc{d}_{i^+}^{s}))}}{\exp{(sim(\textsc{c}_{i}^{s},\textsc{d}_{i^+}^{s}))}~+~\sum_{j=1}^{r} \exp{(sim(\textsc{c}_{i}^{s},\textsc{d}_{j,i^-}^{s}))}},
\end{split}
\end{equation}
where, $n_{s}$ is the number of claims in the source domain, $\exp{(\parm)}$ is the exponential function, $\textsc{c}_{i}^{s}$ is the \textit{i-th} source claim, $\textsc{d}_{i^+}^{s}$ is a relevant evidence document (randomly selected from the set of relevant documents $V_{i}^{s}$), and $r$ is the number of randomly selected irrelevant documents---denoted by $\textsc{d}_{\parm,i^-}^{s}$. If we use stochastic gradient descent for training, we can use the irrelevant in-batch evidence documents as negative examples. The objective term is minimized with respect to the parameters of the two encoders $f_{c}$ and $f_{d}$. To test the model, we can use the similarity between a given claim and all the evidence documents, and then, can return the documents that have the highest similarity score to the claim.

Due to the lack of labeled data, the training algorithm above is not applicable in the target domain. Thus, we propose an approach to exploit unlabeled data to train the claim and document encoders for the target domain. We begin by training a bi-encoder model in the source domain, as explained earlier and shown in Figure \ref{fig:retriever-a}. Then, we freeze the parameters of the source claim encoder, and adversarially \citep{gan,adda} train an encoder in the target domain to mimic the outputs of the source claim encoder, as shown in Figure \ref{fig:retriever-b}. We, then, repeat the same procedure to train a target document encoder by freezing the parameters of the source document encoder---Figure \ref{fig:retriever-c}. Finally, the two adversarially trained target encoders can be used to calculate the similarity between the target claims and the target evidence documents, as illustrated in Figure \ref{fig:retriever-d}.

The objective terms for adversarially training the target claim encoder are:
\begin{equation}
\small
\setlength{\jot}{0pt}
\setlength{\abovedisplayskip}{0pt}
\setlength{\belowdisplayskip}{0pt}
\medmuskip=0mu
\thinmuskip=0mu
\thickmuskip=0mu
\nulldelimiterspace=0pt
\scriptspace=0pt
\begin{split}
\mathcal{L}_{g_c}= &-\mathbb{E}_{\textsc{c}^s{\sim}C^s}\left[ \log~g_c(f_c^s(\textsc{c}^s)) \right]- \\
&\mathbb{E}_{\textsc{c}^t{\sim}C^t}\left[ \log~(1-g_c(f_c^t(\textsc{c}^t))) \right] ,
\end{split}
\end{equation}
and
\begin{equation}
\small
\setlength{\jot}{0pt}
\setlength{\abovedisplayskip}{0pt}
\setlength{\belowdisplayskip}{0pt}
\medmuskip=0mu
\thinmuskip=0mu
\thickmuskip=0mu
\nulldelimiterspace=0pt
\scriptspace=0pt
\begin{split}
\mathcal{L}_{f_c^t}= &\mathbb{E}_{\textsc{c}^t{\sim}C^t}\left[ \log~g_c(f_c^t(\textsc{c}^t)) \right],
\end{split}
\end{equation}
where $g_c$ is the discriminator classifier for the claims, and $f_c^s$ and $f_c^t$ are the source and target claim encoders respectively. The rest of the terms were defined earlier. The two objective terms are minimized with respect to the parameters of $g_c$ and $f_c^t$ respectively. Thus, intuitively, the discriminator learns to distinguish between the claims in the source and target domains, while the target claim encoder gradually learns to produce vectors that are similar to the source vectors. Similarly, we adversarially train the target document encoder as follows:
\begin{equation}
\small
\setlength{\jot}{0pt}
\setlength{\abovedisplayskip}{0pt}
\setlength{\belowdisplayskip}{0pt}
\medmuskip=0mu
\thinmuskip=0mu
\thickmuskip=0mu
\nulldelimiterspace=0pt
\scriptspace=0pt
\begin{split}
\mathcal{L}_{g_d}= &-\mathbb{E}_{\textsc{d}^s{\sim}D^s}\left[ \log~g_d(f_d^s(\textsc{d}^s)) \right]-\\
&\mathbb{E}_{\textsc{d}^t{\sim}D^t}\left[ \log~(1-g_d(f_d^t(\textsc{d}^t))) \right] ,
\end{split}
\end{equation}
and
\begin{equation}
\small
\setlength{\jot}{0pt}
\setlength{\abovedisplayskip}{0pt}
\setlength{\belowdisplayskip}{0pt}
\medmuskip=0mu
\thinmuskip=0mu
\thickmuskip=0mu
\nulldelimiterspace=0pt
\scriptspace=0pt
\begin{split}
\mathcal{L}_{f_d^t}= &\mathbb{E}_{\textsc{d}^t{\sim}D^t}\left[ \log~g_d(f_d^t(\textsc{d}^t)) \right],
\end{split}
\end{equation}
where $g_d$ is the discriminator classifier for the evidence documents, and $f_d^s$ and $f_d^t$ are the source and target document encoders respectively. Note that before training the target encoders, we initialize them with the parameters of the source encoders. During the experiments we observed that this can significantly facilitate their training.

Pretraining encoders has become an integral part of dense retrieval algorithms \citep{dpr,gpl,prompt}. Our approach for training the target claim and document encoders does not impose any restriction on the initialization of the encoders. Therefore, before training the source encoders (Figure \ref{fig:retriever-a}), we use the T5 model \citep{t5} to generate a set of pseudo claims for the unlabeled evidence documents in the target domain. We then use this automatically generated dataset to pretrain a bi-encoder model, to be used in the training algorithm described in this section. See Section \ref{sec:setup} for the implementation details of the pretraining step. In the next section, we discuss our algorithm for training the reader.

\subsection{Representation Alignment for Veracity Prediction} \label{subsec:reader}
To predict the veracity of a given claim, following existing studies \citep{fc-scientific,fc-scientific-zero-shot,fc-survey-new}, we can train a classifier on the concatenation of the corresponding evidence document and the claim---resembling the natural language inference task. If multiple evidence documents exist, we can take the average of the classifier outputs to make the final prediction. In the source domain $S$, where labeled data is available, we can employ this method. However, it is difficult to train such a classifier for the target domain because there is no labeled data in this domain. Thus, we use the retriever that we trained in the previous step, the labeled and unlabeled data in the source domain, and the unlabeled data in the target domain to train such a classifier for the target domain.

We use a distance-based discrepancy reduction loss function to train our model \citep{dan}. Thus, we have:
\begin{equation} \label{eq:alignment}
\small
\setlength{\jot}{0pt}
\setlength{\abovedisplayskip}{0pt}
\setlength{\belowdisplayskip}{0pt}
\medmuskip=0mu
\thinmuskip=0mu
\thickmuskip=0mu
\nulldelimiterspace=0pt
\scriptspace=0pt
\begin{split}
\mathcal{L}= &\frac{1}{n_s}\sum_{i=1}^{n_s} J(\theta(f_r(\textsc{x}_{i}^{s})), y_{i}^{s}) ~ + ~ \lambda \mathcal{D}(f_r(X^{s}), f_r(X^{t})),
\end{split}
\end{equation}
where $J$ is the cross entropy loss, $f_r(\parm)$ is the data encoder, $\theta(\parm)$ is the classifier applied to the output of the encoder, $\textsc{x}_{i}^{s}$ is \textit{i-th} labeled source example, and $X^{s}$ and $X^{t}$ are the sets of unlabeled source and target examples respectively. $\lambda > 0$ is a penalty term. The term $\mathcal{D}$ is the alignment loss, and reduces the discrepancy between the distributions of source and target examples after the encoder layer. We use correlation alignment \citep{coral}, which measures the distance between the second-order statistics of the source and target data. It is defined as follows:
\begin{equation} \label{eq:coral}
\small
\setlength{\jot}{0pt}
\setlength{\abovedisplayskip}{0pt}
\setlength{\belowdisplayskip}{0pt}
\medmuskip=0mu
\thinmuskip=0mu
\thickmuskip=0mu
\nulldelimiterspace=0pt
\scriptspace=0pt
\begin{split}
\mathcal{D} ~=~ &\frac{1}{4 \times d^2} ~~ {\left| M^s  - M^t \right|}^{2}_{F},
\end{split}
\end{equation}
where d is the dimension of the input vectors, and ${\left| \parm \right|}^{2}_{F}$ is the square of Frobeniuns norm. $M^s$ and $M^t$ are the covariance matrices of $f_r(X^{s})$ and $f_r(X^{t})$ respectively. We see that by reducing the distance between the two covariance matrices the discrepancy between the projected representations of the source and target vectors are reduced.

\begin{figure*}[h]
    \centering
    \begin{subfigure}[t]{0.22\linewidth}
        \centering
        \includegraphics[width=\linewidth]{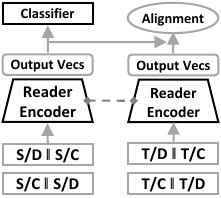}
        \caption{Augmented Reader}
        \label{fig:reader}
    \end{subfigure}~
    \begin{subfigure}[t]{0.35\linewidth}
        \centering
        \includegraphics[width=\linewidth]{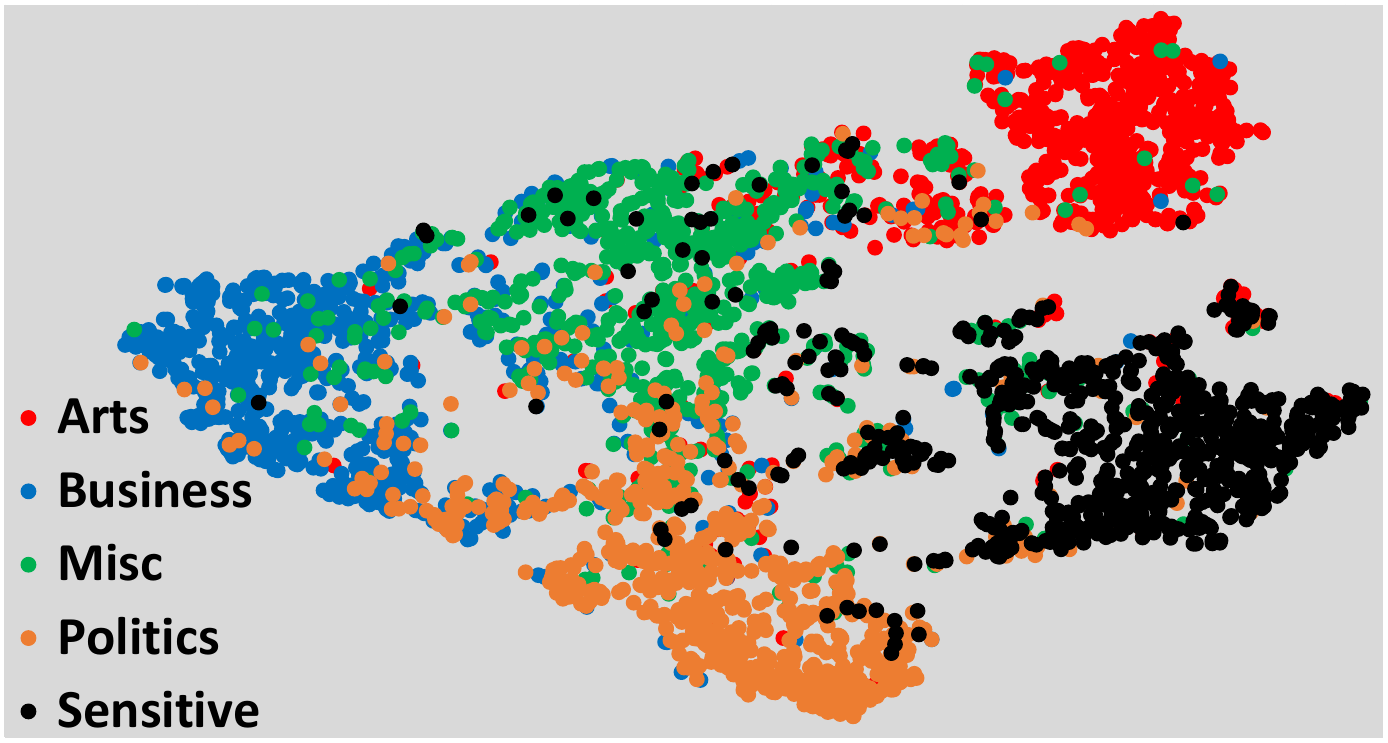}
        \caption{\MULTIFCDS Dataset}
        \label{fig:multifc-scatter}
    \end{subfigure}~
    \begin{subfigure}[t]{0.35\linewidth}
        \centering
        \includegraphics[width=\linewidth]{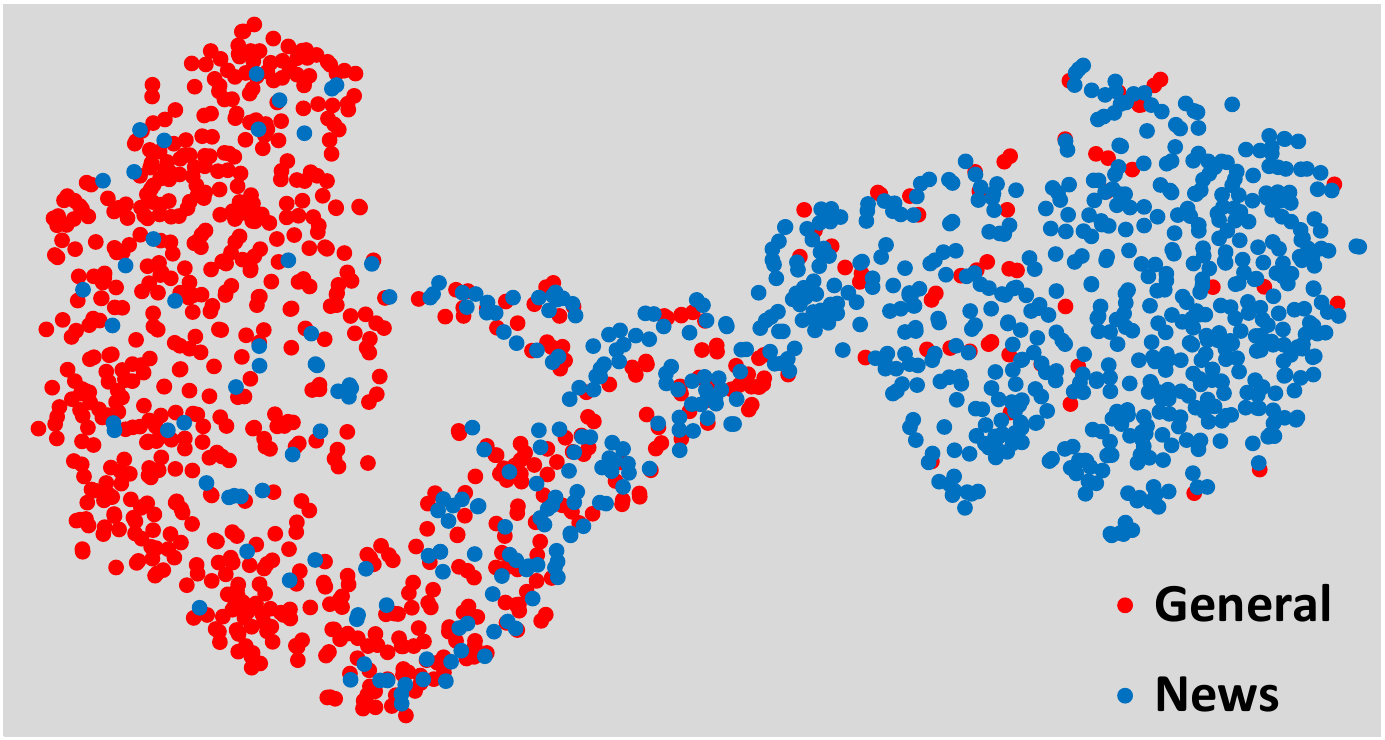}
        \caption{\SNOPESDS Dataset}
        \label{fig:snopes-scatter}
    \end{subfigure}
    \caption{\textbf{\ref{fig:reader})} The reader model. Dashed line indicates shared parameters. S/D, S/C, T/D, and T/C stand for source documents, source claims, target documents, and target claims respectively. The symbol $\mathbin\Vert$ is the concatenation operator. \textbf{\ref{fig:multifc-scatter}-\ref{fig:snopes-scatter})} The 2D projection of the claims in the \MULTIFCDS and \SNOPESDS datasets (using t-SNE). The vectors are the outputs of a BERT classifier, after being trained to predict the domains. Figure best viewed in color.}
\end{figure*}

In order to use Equation \ref{eq:alignment} for training our model, we need to formulate the vectors $\textsc{x}_{\parm}^{s}$, $X^{s}$, and $X^{t}$. We obtain $\textsc{x}_{\parm}^{s}$ and $X^{s}$ in the source domain by concatenating the evidence documents and their corresponding claims. Because there are no associations between the documents and the claims in the target domain, we propose to use the model trained in the previous section to retrieve the top $p$ target documents for each target claim, and then, to consider them as the evidence documents. These documents along their associated claims can be used to construct the vectors $X^{t}$.


To give the reader more cues and also provide it with more training data, we propose to augment the input data with the reverse order of itself. For instance, in the case of $\textsc{x}_{i}^{s}$, if $\textsc{x}_{i}^{s}{=}\textsc{c}_{i}^{s}\mathbin\Vert\textsc{d}_{i^+}^{s}$, where the symbol $\mathbin\Vert$ is the concatenation operator, we then propose to also use the vector $\overline{\textsc{x}_{i}^{s}}{=}\textsc{d}_{i^+}^{s}\mathbin\Vert\textsc{c}_{i}^{s}$ for training the reader. The augmentation can be performed on all the vectors in $X^{s}$ and $X^{t}$ as well. Note that in the general natural language inference task, it is not always logically true to reverse the order of the premise and the hypothesis, however, in the fact checking task this is the case. See Table \ref{tbl:augment} in the results section for an anecdotal experiment that shows a language model fails to infer $B{\rightarrow}A$ from $A{\rightarrow}B$, which justifies our augmentation method.\footnote{Parallel to our paper, another paper by \citet{reversal-curse} reports the same finding.} Note that after the augmentation step, Equation \ref{eq:alignment} will have two discrepancy alignment terms. We use $\lambda_1$ and $\lambda_2$ as coefficients for the direct and reverse alignment terms.


Figure \ref{fig:reader} shows our reader. We see that the input data is augmented with the reverse vectors. The entire model is trained using the supervised cross entropy loss and the unsupervised alignment loss terms.

\subsection{Training and Testing Procedures}
\label{subsec:train-test-proc}

To train our fact checking pipeline, we use the labeled source data and the unlabeled target data in the algorithm presented in Section \ref{subsec:retriever} to train our retriever. Then we use the trained retriever to generate pseudo-labels for the target claims--as mentioned in Section \ref{subsec:retriever}, our target retriever has two encoders adversarially trained for the task. We, then, use the labeled source data along the pseudo-labeled target data to train our reader, as stated in Section \ref{subsec:reader}.

Improving a pipeline by improving each component individually, rather than proposing end-to-end solutions, has a downside.
When our pipeline is used for testing, the improvement achieved by each component may not be fully carried over to the next step. For instance, the retriever may return better results in a particular scenario, but the reader may fail to exploit the informative evidence documents in this scenario. Another example is when the reader can potentially perform better, but the retriever fails to return informative evidence documents. In general, as stated by \citet{domingos}, learning is a complex phenomenon. In order to potentially dampen the undesired effect of such cases, we add an additional step during the testing. During the testing, given an unseen target claim, the retriever is used to return the top $k$ evidence documents, and they are carried over to the reader. At this stage, instead of treating these documents as a set, we use the ranking of the documents to assign a higher weight to the top documents in making the final prediction. Therefore, instead of taking the average of the classifier to derive the prediction, we begin from the top of the list and iterate over the ranking list to generate $k$ subsets. The final prediction is made by taking the average of the predictions obtained from each subset. More formally, the final prediction is made as follows:
\begin{equation}
\small
\setlength{\jot}{0pt}
\setlength{\abovedisplayskip}{0pt}
\setlength{\belowdisplayskip}{0pt}
\medmuskip=0mu
\thinmuskip=0mu
\thickmuskip=0mu
\nulldelimiterspace=0pt
\scriptspace=0pt
\begin{split}
\mathcal{O}= &\frac{1}{k}\sum_{i=1}^{k} (\frac{ \sum_{j=1}^{i} \theta(f_r( \textsc{c}^{t}\mathbin\Vert\textsc{d}_{j^+}^{t} )) }{i}),
\end{split}
\end{equation}
where, as before, $f_r$ and $\theta$ are the reader encoder and the reader classifier, $\textsc{c}^{t}$ is the target claim at hand, and $\textsc{d}_{j^+}^{t}$ is the \textit{j-th} relevant document returned by the retriever. We see that the top evidence documents are present in a higher number of subsets, and therefore, have a higher weight. In the next section, we provide an overview of the experimental setup for evaluating our pipeline.

\begin{table}
\small
\centering
\begingroup 
\setlength{\tabcolsep}{2pt} 
\begin{tabular}{p{0.6in} p{0.5in} p{0.35in} | p{0.45in} p{0.45in} p{0.45in} } \Xhline{3\arrayrulewidth}
\multicolumn{1}{c}{\textbf{Dataset}} & \textbf{Domain} & \textbf{Count} & \textbf{Neutral} & \textbf{Refute} & \textbf{Support} \\ \Xhline{3\arrayrulewidth}
\multicolumn{1}{c}{\multirow{5}{*}{\MULTIFCDS}} & Arts & 3788 & - & 3434 & 354 \\
\multicolumn{1}{c}{} & Business & 1943 & - & 1007 & 936 \\
\multicolumn{1}{c}{} & Misc & 7968 & - & 5351 & 2617 \\
\multicolumn{1}{c}{} & Politics & 9350 & - & 6301 & 3049 \\
\multicolumn{1}{c}{} & Sensitive & 2180 & - & 1555 & 625 \\ \hline
\multicolumn{1}{c}{\multirow{2}{*}{\SNOPESDS}} & General & 4190 & 755 & 2643 & 792 \\
\multicolumn{1}{c}{} & News & 1620 & 348 & 1041 & 231 \\ \Xhline{3\arrayrulewidth}
\end{tabular}
\endgroup 
\caption{The list of domains, the number of claims in each domain, and the distribution of labels in each domain for the \MULTIFCDS and \SNOPESDS datasets.} \label{tbl:dataset-stat}
\end{table}

\begingroup 
\setlength{\tabcolsep}{2.5pt} 
\begin{table*}
\centering
\small
\begin{tabu}{p{0.7in} p{0.43in} p{0.43in} p{0.43in} p{0.43in} p{0.43in} p{0.43in} p{0.43in} p{0.43in} p{0.43in} p{0.43in}} \Xhline{2\arrayrulewidth}
 \cline{1-11} & \multicolumn{7}{c}{\textbf{F1 in \MULTIFCDS}}  &
 \multicolumn{3}{c}{\textbf{F1 in \SNOPESDS}} \\
\cmidrule[\heavyrulewidth](l){2-8} \cmidrule[\heavyrulewidth](l){9-11}
\multicolumn{1}{c}{\textbf{Method}} & 
\textbf{M$\rightarrow$A} & \textbf{M$\rightarrow$B} & \textbf{M$\rightarrow$S} & \textbf{P$\rightarrow$A} 
& \textbf{P$\rightarrow$B} & \textbf{P$\rightarrow$S} & \textbf{Ave} & \textbf{G$\rightarrow$N} & \textbf{N$\rightarrow$G} & \textbf{Ave}
\\ \Xhline{3\arrayrulewidth}
\multicolumn{1}{c}{\textit{cont-gpl-ft/nli-ft}} & 0.580\textsubscript{.02} & 0.593\textsubscript{.01} & 0.638\textsubscript{.01} & 0.579\textsubscript{.02} 
& 0.595\textsubscript{.01} & 0.629\textsubscript{.01} & 0.602 & 0.435\textsubscript{.02} & 0.403\textsubscript{.02} & 0.419 \\
\multicolumn{1}{c}{\textit{cont-gpl-ft/nli-mlm-ft}} & 0.581\textsubscript{.03} & 0.593\textsubscript{.02} & 0.635\textsubscript{.02} & 0.600\textsubscript{.02} & 0.590\textsubscript{.01} & 0.620\textsubscript{.01} & 0.603 & 0.422\textsubscript{.04} & 0.416\textsubscript{.01} & 0.419 \\
\multicolumn{1}{c}{\textit{cont-promp-ft/nli-ft}} & 0.583\textsubscript{.01} & 0.594\textsubscript{.01} & 0.642\textsubscript{.01} & 0.586\textsubscript{.02} 
& \textbf{0.604\textsubscript{.00}} & 0.623\textsubscript{.01} & 0.605 & 0.434\textsubscript{.01} & 0.406\textsubscript{.01} & 0.420 \\
\multicolumn{1}{c}{\textit{cont-promp-ft/nli-mlm-ft}} & 0.589\textsubscript{.03} & 0.594\textsubscript{.02} & 0.638\textsubscript{.02} & 0.603\textsubscript{.02} 
& 0.589\textsubscript{.02} & 0.619\textsubscript{.02} & 0.605 & 0.423\textsubscript{.04} & 0.417\textsubscript{.01} & 0.420 \\ \hline 
\multicolumn{1}{c}{\textit{ours}} & \textbf{0.595\textsubscript{.01}} & \textbf{0.605\textsubscript{.01}} & \textbf{0.648\textsubscript{.01}} & \textbf{0.615\textsubscript{.02}} 
& 0.603\textsubscript{.01} & \textbf{0.643\textsubscript{.01}} & \textbf{0.618} & \textbf{0.440\textsubscript{.02}} & \textbf{0.435\textsubscript{.01}} & \textbf{0.437 }\\ \Xhline{3\arrayrulewidth}
\end{tabu}
\caption{Fact checking results. The sequence before ``/'' indicate the list of steps used in the retriever, and the sequence after ``/'' indicate the list of steps used in the reader. The suffix \textit{ft} indicates finetuning on the source domain. For examples, \textit{cont-promp-ft} means that fist Contriever is used, then Promptagator is used, and finally the model is finetuned on the source domain. For brevity, the initials of the domain names are used in the column titles. All the baselines use domain adaptation techniques. For a comparison to a pipeline that does not use any domain adaptation method see Appendix \ref{sec:appendix-results}.} \label{tbl:fc-results}
\end{table*}
\endgroup 

\begingroup 
\setlength{\tabcolsep}{3pt} 
\begin{table*}
\centering
\small
\begin{tabu}{p{0.7in} p{0.43in} p{0.43in} p{0.43in} p{0.43in} p{0.43in} p{0.43in} p{0.43in} p{0.43in} p{0.43in} p{0.43in}} \Xhline{2\arrayrulewidth}
 \cline{1-11} & \multicolumn{7}{c}{\textbf{F1 in \MULTIFCDS}}  &
 \multicolumn{3}{c}{\textbf{F1 in \SNOPESDS}} \\
\cmidrule[\heavyrulewidth](l){2-8} \cmidrule[\heavyrulewidth](l){9-11}
\multicolumn{1}{c}{\textbf{Method}} & 
\textbf{M$\rightarrow$A} & \textbf{M$\rightarrow$B} & \textbf{M$\rightarrow$S} & \textbf{P$\rightarrow$A} 
& \textbf{P$\rightarrow$B} & \textbf{P$\rightarrow$S} & \textbf{Ave} & \textbf{G$\rightarrow$N} & \textbf{N$\rightarrow$G} & \textbf{Ave}
\\ \Xhline{3\arrayrulewidth}
\multicolumn{1}{c}{\textit{nli}} & 0.443\textsubscript{.06} & 0.446\textsubscript{.06} & 0.451\textsubscript{.04} & 0.443\textsubscript{.06} 
& 0.446\textsubscript{.06} & 0.451\textsubscript{.04} & 0.447 & 0.194\textsubscript{.07} & 0.201\textsubscript{.06} & 0.198 \\
\multicolumn{1}{c}{\textit{nli-ft}} & 0.628\textsubscript{.02} & 0.613\textsubscript{.00} & 0.646\textsubscript{.02} & 0.624\textsubscript{.01} & 0.601\textsubscript{.00} & 0.640\textsubscript{.01} & 0.625 & 0.454\textsubscript{.01} & 0.449\textsubscript{.01} & 0.451 \\
\multicolumn{1}{c}{\textit{nli-mlm-ft}} & 0.614\textsubscript{.00} & 0.611\textsubscript{.01} & 0.648\textsubscript{.02} & 0.629\textsubscript{.03} 
& 0.600\textsubscript{.02} & 0.632\textsubscript{.02} & 0.622 & 0.440\textsubscript{.04} & 0.441\textsubscript{.00} & 0.441 \\ \hline 
\multicolumn{1}{c}{\textit{ours}} & \textbf{0.637\textsubscript{.02}} & \textbf{0.625\textsubscript{.01}} & \textbf{0.662\textsubscript{.01}} & \textbf{0.639\textsubscript{.01}} 
& \textbf{0.611\textsubscript{.02}} & \textbf{0.651\textsubscript{.01}} & \textbf{0.637} & \textbf{0.466\textsubscript{.01}} & \textbf{0.469\textsubscript{.01}} & \textbf{0.467 }\\ \Xhline{3\arrayrulewidth}
\end{tabu}
\caption{The performance of the reader compared to the baselines. The suffix \textit{ft} indicates finetuning on the source domain.} \label{tbl:reader-results}
\end{table*}
\endgroup 

\section{Experimental Setup} \label{sec:setup}

We begin this section by providing an overview of the datasets used in the experiments. Afterwards, we briefly discuss the baselines that we compare to, and finally, we present a summary of the setup. Additional information about the baselines and the training setup can be found in Appendix.

\noindent\textbf{Datasets.} We use two datasets in our experiments, the \MULTIFCDS dataset \citep{multi-fc-ds} and the \SNOPESDS dataset \citep{snopes-ds}. The claims in these datasets are not categorized into domains, therefore, we automatically extract the domains. See Appendix \ref{sec:appendix-datasets} for a description about the process, a sample set of the claims from each domain, and the top LDA topics of the domains. Table~\ref{tbl:dataset-stat} reports the list of the domains, and the distribution of the labels in each domain. We also report the 2D projections of the claim representations in Figures \ref{fig:multifc-scatter} and \ref{fig:snopes-scatter}. These illustrations are the outputs of a BERT encoder trained to project the claim representations, then further transformed into 2D vectors using the t-SNE technique \citep{t-sne}. We observe that there is a marked shift between the distributions of each pair of the domains in both datasets.

\noindent\textbf{Baselines.} A detailed description of each baseline and the setup can be found in Appendix \ref{sec:appendix-setup}. We compare our retriever with three baselines \citet{contriever}, \citet{gpl}, and \citet{prompt}. The former of the list is a pretraining technique. To evaluate our reader, we show that it outperforms a commonly used domain adaptation method called DAPT \citep{do-pretrain} followed by finetuning in the source domain. In Table \ref{tbl:ablation-reader} we also compare our augmentation model to the vanilla distance-based domain adaptation method.

We compare our fact checking pipeline to a set of pipelines that consist of the best retriever components cross connected to the best reader components.

\noindent\textbf{Setup.} We follow the standard practice in domain adaptation literature \citep{dom-ada-theory} to carry out the experiments. We take several domains as source and the rest as the target domains. During training we assume we don't have access to the target labels, and use them only for testing. In the \MULTIFCDS dataset, we use the domains Misc and Politics as the source and the rest as target. We select these two as source domains because they have the highest A-distance from the rest of the domains-with 0.09 and 0.07 respectively, compared to 0.06, 0.05 and 0.04 for Business, Sensitive, and Arts respectively.\footnote{A-distance \citep{dom-ada-theory} is a measure of discrepancy between two domains, and can be approximated by the error rate of a classifier trained to labels the samples from the two domains \citep{a-distance-approx}.} In \SNOPESDS\!, we use both domains iteratively as source and target.


\begingroup 
\setlength{\tabcolsep}{3pt} 
\begin{table*}
\centering
\small
\begin{tabu}{p{0.7in} p{0.43in} p{0.43in} p{0.43in} p{0.43in} p{0.43in} p{0.43in} p{0.43in} p{0.43in} p{0.43in} p{0.43in}} \Xhline{2\arrayrulewidth}
 \cline{1-11} & \multicolumn{7}{c}{\textbf{NDCG@10 in \MULTIFCDS}}  &
 \multicolumn{3}{c}{\textbf{NDCG@10 in \SNOPESDS}} \\
\cmidrule[\heavyrulewidth](l){2-8} \cmidrule[\heavyrulewidth](l){9-11}
\multicolumn{1}{c}{\textbf{Method}} & 
\textbf{M$\rightarrow$A} & \textbf{M$\rightarrow$B} & \textbf{M$\rightarrow$S} & \textbf{P$\rightarrow$A} 
& \textbf{P$\rightarrow$B} & \textbf{P$\rightarrow$S} & \textbf{Ave} & \textbf{G$\rightarrow$N} & \textbf{N$\rightarrow$G} & \textbf{Ave}
\\ \Xhline{3\arrayrulewidth}
\multicolumn{1}{c}{\textit{bm25}} & 0.684\textsubscript{} & 0.723\textsubscript{} & 0.725\textsubscript{} & 0.684\textsubscript{} & 0.723\textsubscript{} & 0.725\textsubscript{} & 0.711 & 0.558\textsubscript{} & 0.638\textsubscript{} & 0.598 \\
\multicolumn{1}{c}{\textit{cont-ft}} & 0.673\textsubscript{.01} & 0.654\textsubscript{.01} & 0.707\textsubscript{.00} & 0.700\textsubscript{.00} & 0.663\textsubscript{.01} & 0.714\textsubscript{.01} & 0.685 & 0.577\textsubscript{.01} & 0.737\textsubscript{.00} & 0.657 \\
\multicolumn{1}{c}{\textit{cont-t5}} & 0.721\textsubscript{.00} & 0.624\textsubscript{.00} & 0.711\textsubscript{.00} & 0.721\textsubscript{.00} & 0.624\textsubscript{.00} & 0.711\textsubscript{.00} & 0.685 & 0.623\textsubscript{.00} & 0.737\textsubscript{.00} & 0.680 \\
\multicolumn{1}{c}{\textit{cont-gpl-ft}} & 0.794\textsubscript{.00} & 0.734\textsubscript{.01} & 0.784\textsubscript{.00} & 0.801\textsubscript{.00} & 0.748\textsubscript{.00} & 0.788\textsubscript{.00} & 0.774 & 0.642\textsubscript{.00} & 0.769\textsubscript{.00} & 0.705 \\
\multicolumn{1}{c}{\textit{cont-promp-ft}} & 0.785\textsubscript{.00} & 0.723\textsubscript{.00} & 0.773\textsubscript{.01} & 0.796\textsubscript{.00} 
& 0.735\textsubscript{.01} & 0.776\textsubscript{.00} & 0.764 & 0.637\textsubscript{.00} & 0.766\textsubscript{.00} & 0.702 \\ \hline 
\multicolumn{1}{c}{\textit{ours}} & \textbf{0.803\textsubscript{.00}} & \textbf{0.747\textsubscript{.01}} & \textbf{0.793\textsubscript{.00}} & \textbf{0.810\textsubscript{.00}} 
& \textbf{0.757\textsubscript{.00}} & \textbf{0.797\textsubscript{.00}} & \textbf{0.784} & \textbf{0.647\textsubscript{.00}} & \textbf{0.773\textsubscript{.00}} & \textbf{0.710} \\ \Xhline{3\arrayrulewidth}
\end{tabu}
\caption{The performance of the retriever compared to the baseline models. The suffix \textit{ft} indicates finetuning on the source domain. The suffix \textit{t5} indicates finetuning on synthetically generated T5 queries.} \label{tbl:retriever-results}
\end{table*}
\endgroup 

\begingroup 
\setlength{\tabcolsep}{3pt}
\begin{table*}
\small
  \centering
  \subfloat[Retriever Ablation Studies]{
  \begin{tabular}{p{0.7in} p{0.4in} p{0.4in}} \Xhline{2\arrayrulewidth}
    \toprule
    \textbf{Method} & P$\rightarrow$S & N$\rightarrow$G \\
    \midrule
    Full & 0.797 & 0.773 \\
    w/o claim enc & 0.778 & 0.770 \\
    w/o doc enc & 0.792 & 0.769 \\ \Xhline{3\arrayrulewidth}
  \end{tabular}
  \label{tbl:ablation-retriever}
  }~
  \subfloat[Reader Ablation Studies]{
    \begin{tabular}{p{0.65in} p{0.4in} p{0.4in}} \Xhline{2\arrayrulewidth}
    \toprule
    \textbf{Method} & P$\rightarrow$S & N$\rightarrow$G \\
    \midrule
    Full & 0.651 & 0.469 \\
    w/o align & 0.638 & 0.464 \\
    w/o reverse & 0.646 & 0.455 \\ \Xhline{3\arrayrulewidth}
  \end{tabular}
  \label{tbl:ablation-reader}
  }~
  \subfloat[Pipeline Ablation Studies]{
    \begin{tabular}{p{0.65in} p{0.4in} p{0.4in}} \Xhline{2\arrayrulewidth}
    \toprule
    \textbf{Method} & P$\rightarrow$S & N$\rightarrow$G \\
    \midrule
    Full & 0.643 & 0.435 \\
    w/o retriever & 0.640 & 0.432 \\
    w/o reader & 0.632 & 0.404 \\
    w/o ranking & 0.636 & 0.413 \\ \Xhline{3\arrayrulewidth}
  \end{tabular}
  \label{tbl:ablation-fc}
  }
  \caption{Ablation studies of the proposed methods in the retriever~(\ref{tbl:ablation-retriever}), the reader~(\ref{tbl:ablation-reader}), and pipeline~(\ref{tbl:ablation-fc}) for a use case in the \MULTIFCDS dataset (P$\rightarrow$S) and in the \SNOPESDS dataset (N$\rightarrow$G) .}%
\end{table*}
\endgroup 

\section{Results and Analysis} \label{sec:results}

\begin{table*}
\aboverulesep=0ex
\belowrulesep=0ex
\centering
\begin{tabular}{p{1.0in} | p{3.5in} } \Xhline{3\arrayrulewidth}
\textbf{Set} & \textbf{Prompt} \\ \Xhline{3\arrayrulewidth}
Train & \makecell[l]{Evidence: “MIT is the alma mater of GHI.” \\ Claim: “GHI studied at MIT.” \\ Label: “Is that true or false? True”} \\
\cmidrule(l){1-2}
Test & \makecell[l]{Evidence: “ABC studied at University of Illinois.” \\ Claim: “University of Illinois is the alma mater of ABC.” \\ Label: “Is that true or false?”} \\
\cmidrule(l){1-2}
Augmentation & \makecell[l]{Evidence: “GHI studied at MIT.” \\ Claim: “MIT is the alma mater of GHI.” \\ Label: “Is that true or false? True”} \\ \Xhline{3\arrayrulewidth}
\end{tabular}
\caption{An example that GPT-3 fails to infer the reversal relationship between the evidence and claim. If only the train and test rows are used in the prompt, the model fails to output the correct answer---the correct answer is True. However, if the prompt is augmented with the reverse of the train row, then the model outputs the correct answer.} \label{tbl:augment}
\end{table*}

Table \ref{tbl:fc-results} reports the results of the fact checking pipeline across the two datasets for our model compared to the baseline methods. We observe that in all the scenarios our model is either the top performing approach, or is on a par with the best method. 

In Tables \ref{tbl:reader-results}, we report the performance of the reader compared to the alternative methods individually. To evaluate the reader in isolation, we assume that the retriever returns all the relevant evidence documents. The first observation is that our model is able to offer a lot of improvement on top of the \textit{nli} pretraining model--this model is pretrained on SNLI and MultiNLI datasets. All the methods (including ours) uses this model as the starting checkpoint for training. We also observe that the gap between our model and the baselines is still present even if we finetune the pretrained model on the source dataset (\textit{nli-ft}). This does not change even if we pretrain the model on the target domain using the masked language model task~(\textit{nli-mlm-ft}).

In Table \ref{tbl:retriever-results}, we report the performance of our retriever compared to the baselines individually.\footnote{In a separate experiment, we tried to visualize the embedding space of the representations before and after the adaptation. However, we found that it is difficult to qualitatively observe the improvements. Thus, we resort to quantitative evaluations.} Again we see that our retriever outperforms the basic BM25 model and the pretraining model finetuned on the source data (\textit{cont-ft}) by a large margin. We don't report the plain Contriever model as it was unable to solve the task reasonably. However, all the methods (including ours) use this model as the starting point for training. A noteworthy observation from Table \ref{tbl:retriever-results} is that the performance of \textit{cont-ft} and \textit{cont-t5} is on a par with each other. One of them (\textit{cont-ft}) is only finetuned on the source data, and the other one (\textit{cont-t5}) is only finetuned on synthetically generated target data using T5. 

To better understand the properties of our model, we report a series of ablation studies in both components of the pipeline, as well as the entire pipeline itself. In Table \ref{tbl:ablation-retriever}, we report the performance when we omit the adversarial training of the encoders individually. We observe that each step is relatively contributing to the results. In Table \ref{tbl:ablation-reader}, we repeat the same experiment by omitting the alignment loss term and the reversal augmentation. We see that both steps are noticeably enhancing the performance. Finally, to evaluate the components within the pipeline, in Table \ref{tbl:ablation-fc}, we report the performance when we disable our algorithms in the retriever, in the reader, and in ranking the top evidence documents. We see that each component is relatively boosting the performance, however, as stated by \citet{fc-survey-new}, even though the retriever individually shows improvement, when it is within the pipeline it demonstrates less effectiveness.

In Table \ref{tbl:ablation-reader}, we quantitatively show that it is an effective strategy to augment the input data with the reverse order of itself for training the reader. It is informative to see if this strategy can still be helpful if used along existing large language models. To this end, we report an experiment in Table \ref{tbl:augment}, where we use GPT-3 to validate a claim given an evidence document. We see that if we only use the direct association between the claim and the evidence for in-context learning, the model fails to answer a similar question. However, if we augment the input data with the reverse data point, the model can make the right choice.

\section{Conclusions} \label{sec:conclusions}

We studied automatic fact checking under domain shift. We showed that large language models are unable to do the task in certain cases. Then we empirically showed that the common fact checking pipeline suffers from distribution shift, when it is trained in one domain and tested in another domain. We then proposed two novel algorithms to enhance the performance of the entire pipeline. We evaluated our model in eight scenarios and showed that in the majority of the cases our model is the top performing algorithm.

\section*{Acknowledgements}

This research is based upon work supported by U.S. DARPA SemaFor Program No. HR001120C0123 and DARPA KAIROS Program No. FA8750-19-2- 1004. The views and conclusions contained herein are those of the authors and should not be interpreted as necessarily representing the official policies, either expressed or implied, of DARPA, or the U.S. Government. The U.S. Government is authorized to reproduce and distribute reprints for governmental purposes notwithstanding any copyright annotation therein. 

We thank the anonymous reviewers for their feedback.

\section{Limitations} \label{sec:limit}

First limitation of our study is that it focuses only on textual data. Fact verification can be also performed on knowledge graphs. We selected textual data due to the popularity of this type of knowledge source. Second limitation of our study is that it only reports experiments in English language. This was imposed on us due to the lack of large-scale fact checking datasets in other languages. The third limitation, which is connected to the previous shortcoming, is the lack of multiple domain benchmark for fact checking. 
We acknowledge that our work could be improved by manually composing a large-scale multiple domain fact checking dataset. One potential solution that we considered was to run experiments across multiple datasets. However, as stated by \citet{label-shift}, this introduces another technical challenge called label-shift, which was out of scope of our study.

\bibliography{custom}

\appendix

\renewcommand{\arraystretch}{1.2}

\begin{table*}
\aboverulesep=0ex
\belowrulesep=0ex
\centering
\begin{tabular}{p{0.7in} p{1.1in} | p{4.0in} } \Xhline{3\arrayrulewidth}
\multicolumn{1}{c}{\textbf{Dataset}} & \textbf{Mapped Domain} & \textbf{Google Content Classification Labels} \\ \Xhline{3\arrayrulewidth}
\multicolumn{1}{c}{\multirow{7}{*}{MultiFC}} & \multicolumn{1}{c|}{\multirow{1}{*}{Arts}} &  /Arts \& Entertainment \\
\cmidrule(l){2-3}
\multicolumn{1}{c}{} & \multicolumn{1}{c|}{\multirow{3}{*}{Business}} & /Finance \newline
/Business \newline
/News/Business News \\
\cmidrule(l){2-3}
\multicolumn{1}{c}{} & \multicolumn{1}{c|}{\multirow{2}{*}{Politics}} & /Law \& Government \newline
/News/Politics \\ 
\cmidrule(l){2-3}
\multicolumn{1}{c}{} & \multicolumn{1}{c|}{\multirow{1}{*}{Sensitive}} & /Sensitive Subjects \\
\cmidrule(l){2-3}
\multicolumn{1}{c}{} & \multicolumn{1}{c|}{\multirow{1}{*}{Misc}} & The Rest Of The Labels \\ \Xhline{3\arrayrulewidth}
\multicolumn{1}{c}{\multirow{5}{*}{Snopes}} & \multicolumn{1}{c|}{\multirow{5}{*}{News}} & /News/Politics/Other \newline
/News/Politics/Campaigns \& Elections \newline
/Law \& Government/Government/Executive Branch \newline
/Law \& Government/Public Safety/Crime \& Justice \newline
/News/Other \\
\cmidrule(l){2-3}
\multicolumn{1}{c}{} & \multicolumn{1}{c|}{\multirow{1}{*}{General}} & The Rest Of The Labels \\ \Xhline{3\arrayrulewidth}
\end{tabular}
\caption{The chart used for mapping the Google content classification labels to the domain names in each dataset.} \label{tbl:label-mapping}
\end{table*}

\begin{table*}[h]
\aboverulesep=0ex
\belowrulesep=0ex
\centering
\begin{tabular}{p{0.7in} p{0.7in} | p{3.5in} } \Xhline{3\arrayrulewidth}
\multicolumn{1}{c}{\textbf{Dataset}} & \textbf{Domain} & \textbf{Claim Example} \\ \Xhline{3\arrayrulewidth}
\multicolumn{1}{c}{\multirow{8}{*}{MultiFC}} & Arts & Jennifer Lopez, Alex Rodriguez Marrying In The Spring?  \\
\cmidrule(l){2-3}
\multicolumn{1}{c}{} & Business & For the first time in history the North Atlantic is empty of cargo ships in-transit \\
\cmidrule(l){2-3}
\multicolumn{1}{c}{} & Misc & Samuel Adams Set to Release New Helium Beer \\
\cmidrule(l){2-3}
\multicolumn{1}{c}{} & Politics & Hillary Clinton wore a secret earpiece during the first presidential debate of 2016 \\ 
\cmidrule(l){2-3}
\multicolumn{1}{c}{} & Sensitive & A man died in a meth lab explosion after attempting to light his own flatulence \\ \Xhline{3\arrayrulewidth}
\multicolumn{1}{c}{\multirow{4}{*}{Snopes}} & General & The modern image of Santa Claus was created by the Coca-Cola Company \\
\cmidrule(l){2-3}
\multicolumn{1}{c}{} & News & Donald Trump personally sent out an airplane to bring home U.S. military members stranded in Florida \\ \Xhline{3\arrayrulewidth}
\end{tabular}
\caption{Randomly selected claims from each domain of the MultiFc and Snopes datasets.} \label{tbl:random-claims}
\end{table*}

\begin{table*}[h]
\aboverulesep=0ex
\belowrulesep=0ex
\centering
\begin{tabular}{p{0.7in} p{0.7in} | p{3.5in} } \Xhline{3\arrayrulewidth}
\multicolumn{1}{c}{\textbf{Dataset}} & \textbf{Domain} & \textbf{Most Probable Topic} \\ \Xhline{3\arrayrulewidth}
\multicolumn{1}{c}{\multirow{8}{*}{MultiFC}} & Arts & \makecell[l]{fight, Matthew, Sarah, Jessica, Parker \\ Perry, Katy, Bloom, Orlando, Smith} \\
\cmidrule(l){2-3}
\multicolumn{1}{c}{} & Business & \makecell[l]{tax, home, state, \$, trust \\ pension, fund, work, one, say} \\
\cmidrule(l){2-3}
\multicolumn{1}{c}{} & Misc & \makecell[l]{page, prayer, base, Disney, elect \\ turn, charge, improve, form, 2015} \\
\cmidrule(l){2-3}
\multicolumn{1}{c}{} & Politics & \makecell[l]{Meghan, Markle, Prince, Governor, political \\ public, day, school, record, voting} \\ 
\cmidrule(l){2-3}
\multicolumn{1}{c}{} & Sensitive & \makecell[l]{Shooting, wear, agree, Pat \\ involve, media, crash, car, send} \\ \Xhline{3\arrayrulewidth}
\multicolumn{1}{c}{\multirow{3}{*}{Snopes}} & General & \makecell[l]{announce, plan, California, group, Airline \\ document, series, Google, movie, mosque} \\
\cmidrule(l){2-3}
\multicolumn{1}{c}{} & News & \makecell[l]{Donald, Trump, use, U.S., President \\ Clinton, Hillary, e-mail, WikiLeaks, Trump} \\ \Xhline{3\arrayrulewidth}
\end{tabular}
\caption{Top two LDA topics for each domain of the MultiFc and Snopes datasets.} \label{tbl:lda-topics}
\end{table*}

\section{Complementary Reports About the Datasets} \label{sec:appendix-datasets}

We use two datasets in our experiments, the \MULTIFCDS dataset \citep{multi-fc-ds} and the \SNOPESDS dataset \citep{snopes-ds}. The claims in these datasets are not categorized into domains, therefore, we propose a straightforward method to automatically assign a domain name to each claim. To do so, we employ a general purpose classifier trained on a large set of categories. We opt for using the Google Content Classifier,\footnote{Available at: https://cloud.google.com/natural-language/docs/classify-text-tutorial} which is a multi-class model with 1,091 class labels. The labels assigned by the Google API are fine-grained, and in some cases, semantically close to each other. Therefore, we use a manually-crafted chart to map the Google labels to domain names. We constructed five domains in \MULTIFCDS dataset and two domains in \SNOPESDS dataset. Table \ref{tbl:dataset-stat} reports a summary of the domains, and the distribution of the labels in each domain. The claims in the \SNOPESDS dataset are categorized into three veracity labels, whereas, the claims in the \MULTIFCDS dataset cover a much wider range of 179 labels. Due to the nature of this dataset, in many cases the labels are not easily interpretable. To make this dataset suitable for the regular fact checking task, we assign the label ``Support'' to those claims that are labeled as ``True'', and consider the rest as ``Refute''. Additionally, the evidence documents in the \MULTIFCDS dataset are collected through the Google search engine. In the majority of the cases that we inspected, we found that the evidence documents either all support or all refute their associated claims. We conjecture that the Google search engine internally verifies user claims, and retrieves consistent evidence documents, rather than retrieving potentially conflicting information. To have a more realistic evaluation setting, for each claim in this dataset, we randomly selected two evidence documents and discarded the rest of them. We make all the claims, along their domain names, and their labels publicly available for full reproducibility. 

In Table \ref{tbl:label-mapping}, we report the chart that we used to map the Google labels to the domain names in \MULTIFCDS and \SNOPESDS datasets. In Table \ref{tbl:random-claims}, we report a set of randomly selected claims from each domain of the two datasets. Table \ref{tbl:lda-topics} reports the top two topics extracted from the claims of each domain using the Latent Dirichlet Allocation algorithm \citep{lda}.

\section{Complementary Information About the Training Setup} \label{sec:appendix-setup}

\noindent\textbf{Baselines.} We evaluate our retriever from two aspects: first, we show that it is able to offer improvement over common pretraining techniques in dense retrieval for domain adaptation, and second, we show that it outperforms state-of-the-art dense retrieval methods for domain adaptation in the fact checking task. As the pretraining method, we use the model proposed by \citet{contriever}, called Contriever. This model is an unsupervised method based on contrastive learning by cropping spans of texts from documents and taking them as positive samples.  Additionally, we compare to the models proposed by \citet{gpl} and \citet{prompt}, called GPL and Promptagator. GPL uses a query generator, pretrained on the MSMarco dataset, to generate pseudo-queries for the target documents. These pseudo-queries are used to pretrain the dense retrieval model. Promptagator, is a prompt-based model that uses a large language model to generate pseudo-queries for the target documents to be used for finetuning. To have a fair comparison between the models, all of them employ an identical underlying architecture (a bi-encoder) and pre-training steps (using Contriever). The encoder in Contriever is a BERT-sized transformer-based language model, which is used in all the models. Promptagator uses a large language model for generating pseudo-labels. We use GPT 4 to generate this data. We follow the instructions stated by \citet{prompt} and generate 5,000 pseudo-labels for each domain, to be used for pretraining in this model. In addition to these baseline models, we also compare our model to the traditional BM25 model. 

We follow the same protocol for evaluating the reader. We show that it is able to offer improvement over a relevant general domain pretraining task. For this purpose, we use the Roberta model (base variant) \citep{roberta} pretrained on two NLI datasets, i.e., SNLI and MultiNLI datasets \citep{multi-nli}. Then, we also show that it outperforms a common model proposed by existing literature, which is pretraining on the masked language model task (mlm) in the target domain, and then, finetuning in the source domain. To evaluate the entire fact checking pipeline, we compare our model to the pipelines that are constructed by cross connecting two top retrievers to two top readers.

\noindent\textbf{Setup.} Our model has a few hyper-parameters. One for the coefficient of the alignment loss, and another one for the coefficient of the reverse terms--both subjects were discussed in Section \ref{subsec:reader}. We used the domains Misc and Politics in \MULTIFCDS, and searched for the best values between \{0.1,0.3,0.5,0.7,0.9\}. The best values for both is 0.1. We set the value of $K$ in the reader to 10 across all the experiments--$K$ is the top documents returned by the retriever. As the alignment loss term--introduced in Equation \ref{eq:alignment}, we use a metric called correlation alignment \citep{coral}, which measures the distance between the second-order statistics of the source and target data. For pretraining our retriever, we use a T5 model trained on the MS-Marco dataset and generate 3 pseudo claims for each evidence document and pretrain the retriever for three epochs. We set the batch size in the retriever to 70, and in the reader to 50. We set the max sequence size for the claims to 50, and for the documents to 200. We use Adam optimizer in all the experiments. We also use gradient check-pointing for compression. We repeat all the experiments five times, and report the average results. We used four NVIDIA Tesla V100 GPUs with 16G of RAM to run our experiments. The experiments took less than one month to finish.

\begingroup 
\setlength{\tabcolsep}{3pt} 
\begin{table*}
\centering
\small
\begin{tabu}{p{0.7in} p{0.43in} p{0.43in} p{0.43in} p{0.43in} p{0.43in} p{0.43in} p{0.43in} p{0.43in}} \Xhline{2\arrayrulewidth}
 \cline{1-9} & \multicolumn{6}{c}{\textbf{F1 in \MULTIFCDS}}  &
 \multicolumn{2}{c}{\textbf{F1 in \SNOPESDS}} \\
\cmidrule[\heavyrulewidth](l){2-7} \cmidrule[\heavyrulewidth](l){8-9}
\multicolumn{1}{c}{\textbf{Method}} & 
\textbf{M$\rightarrow$A} & \textbf{M$\rightarrow$B} & \textbf{M$\rightarrow$S} & \textbf{P$\rightarrow$A} 
& \textbf{P$\rightarrow$B} & \textbf{P$\rightarrow$S} & \textbf{G$\rightarrow$N} & \textbf{N$\rightarrow$G} 
\\ \Xhline{3\arrayrulewidth}
\multicolumn{1}{c}{\textit{W/O DA}} & 0.590 & 0.583 & 0.607 & 0.610 & 0.573 & 0.605 & 0.383 & 0.391 \\
\multicolumn{1}{c}{\textit{ours}} & \textbf{0.595} & \textbf{0.605} & \textbf{0.648} & \textbf{0.615} & \textbf{0.603} & \textbf{0.643} & \textbf{0.440} & \textbf{0.435} \\ \Xhline{3\arrayrulewidth}
\end{tabu}
\caption{Comparison between our model and a pipeline that does not employ domain adaptation techniques.} \label{tbl:no-da}
\end{table*}
\endgroup 

\begingroup 
\setlength{\tabcolsep}{3pt} 
\begin{table*}
\centering
\small
\begin{tabu}{p{0.7in} p{0.43in} p{0.43in} p{0.43in} p{0.43in} p{0.43in} p{0.43in} p{0.43in} p{0.43in}} \Xhline{2\arrayrulewidth}
 \cline{1-9} & \multicolumn{6}{c}{\textbf{F1 in \MULTIFCDS}}  &
 \multicolumn{2}{c}{\textbf{F1 in \SNOPESDS}} \\
\cmidrule[\heavyrulewidth](l){2-7} \cmidrule[\heavyrulewidth](l){8-9}
\multicolumn{1}{c}{\textbf{Method}} & 
\textbf{M$\rightarrow$A} & \textbf{M$\rightarrow$B} & \textbf{M$\rightarrow$S} & \textbf{P$\rightarrow$A} 
& \textbf{P$\rightarrow$B} & \textbf{P$\rightarrow$S} & \textbf{G$\rightarrow$N} & \textbf{N$\rightarrow$G} 
\\ \Xhline{3\arrayrulewidth}
\multicolumn{1}{c}{\textit{GPT-3}} & 0.456 & 0.536 & 0.530 & 0.456 & 0.536 & 0.530 & 0.302 & 0.304 \\
\multicolumn{1}{c}{\textit{ours}} & \textbf{0.595} & \textbf{0.605} & \textbf{0.648} & \textbf{0.615} & \textbf{0.603} & \textbf{0.643} & \textbf{0.440} & \textbf{0.435} \\ \Xhline{3\arrayrulewidth}
\end{tabu}
\caption{Comparison between our model and GPT-3. We use in-context learning to obtain the results of GPT-3. For each label in the datasets, we use two randomly selected claims along with one evidence document for each one as the in-context examples. This results in four examples in the MultiFc dataset, and six examples in the Snopes dataset. We instruct the model to return the exact labels. In the cases that the returned string is not interpretable, we assume the claim is categorized as false.} \label{tbl:gpt3}
\end{table*}
\endgroup 

\section{Complementary Related Work} \label{sec:appendix-rel-work}


There exist a few studies that investigate the transferability of the fact checking models across fact checking websites \citep{multi-fc-ds,fc-scientific,fc-multi-ling}. \citet{multi-fc-ds} compose a data set called MultiFC. This dataset was collected across multiple fact checking websites, which the authors call them “sources/domains”. Their model is the standard retriever-reader pipeline, and their experiments are carried out within each website individually. Their model relies on meta-data collected from webpages. They propose no algorithm for training a model on one domain and testing on another domain. \citet{fc-scientific} compose a dataset called SciFact, collected from scientific repositories. Their model is the standard retriever-reader. To evaluate the tranferibility of their pipeline, they pretrain the pipeline on the claims extracted from wikipedia and then test it on their dataset. Thus, their solution for domain adaptation is to pretrain the pipeline on one resource and then test it on another resource; beyond this, they propose no domain adaptation method. Their study also has a shortcoming: the wikipedia claims that they use to pretrain their pipeline, may share some knowledge with the claims in their dataset. This can potentially distort their conclusions. \citet{fc-multi-ling} compose a multilingual fact checking dataset. This dataset consists of claims, and evidence documents retrieved from Google. They use the standard pipeline, and similar to the second study, they evaluate the transferability of their pipeline by training on the data from one website and testing it on another website. Beyond this, they propose no additional solution for domain adaptation. As opposed to these studies, we delve into the two primary components of the fact checking pipeline, i.e., the retriever and the reader, and propose algorithms to enhance their robustness. Furthermore, to evaluate our model, we do not rely on comparing the results across fact checking websites, instead, we evaluate the transferability across genres of claims. 

Automatic fact checking is a very active research area, interested reader can see numerous surveys published in recent years, such as the works by \citet{fc-survey-1}, \citet{fc-survey-2}, \citet{fc-survey-3}, \citet{fc-survey}, \citet{fc-survey-4}, and \citet{fc-survey-new}. In this study, our goal is not to present an overview of existing fact checking methods, but to focus on a rather unexplored aspect of this subject, i.e., the transferability of common fact checking tools across domains. Previous studies focus on other aspects of the fact checking pipeline. For instance, \citet{fc-graph-evidence-1} and \citet{fc-graph-evidence-2} exploit the unstructured nature of the evidence documents and propose to use graph networks for modeling the relationship between the documents. \citet{fc-seq-2-seq} concatenate all the evidence documents and use a T5 network to model the final step in the pipeline as a sequence-to-sequence problem. They report that introducing noise to the training of T5 enhances the robustness of the pipeline. \citet{fc-generative-retrieve} enhance the first component of the fact checking pipeline---i.e., the retriever---by proposing a generative model to produce document titles (instead of retrieving them) to be used for retrieving evidence sentences. 

There are also an overwhelming number of studies on dense text retrieval published in recent years, see the surveys by \citet{ir-survey-1} and \citet{ir-survey-2}. The method proposed by \citet{ir-adv} relies on a model called domain classifier to push the representations of source and target data points close to each other. However, as they state, because the transformation happens concurrently to the training of the retrieval encoders, it causes instability in the training. Therefore, they cache the representations of the vectors in the previous steps, and include them in their loss function. The most promising methods for domain adaptation in recent years have been those based on pseudo-query generation, such as the methods by \citet{gpl} and \citet{prompt}. The first method \citep{gpl} uses a pretrained model to generate pseudo-queries in the target domain. The second study \citep{prompt} uses a large language model for achieving the same goal.

\section{Complementary Experiments} \label{sec:appendix-results}

In this section, we report two complementary experiments. 
First, we report a comparison between our method and a fact checking pipeline that does not use any domain adaptation technique. This model is finetuned on the source domain, and then, tested on the target domain. Table \ref{tbl:no-da} reports the results. We observe that in all the scenarios our model outperforms the mentioned baseline model, in some cases such as M$\rightarrow$S and G$\rightarrow$N by a large margin.

Second, we report a comparison between our model and GPT-3. In Section \ref{sec:intro}, we empirically showed that large language models are not suitable for every day fact checking tasks, because their corpus is not regularly updated. However, it is still informative to see how these models perform on our datasets. Please note that a direct comparison between our model and a large language model is not fair, because our model requires much less hardware than these models. On the other hand, one may argue that our model has access to evidence documents. Nevertheless, given the large pretraining corpus of these models, it is also very likely that these models are pretrained directly on fact checking websites. This means that they may already have access to the ground-truth labels of the datasets in their parametric knowledge. Considering all these caveats, we report the comparison in Table \ref{tbl:gpt3}.

\end{document}